\pdfoutput=1

\documentclass[11pt]{article}

\usepackage[final]{acl}

\usepackage{times}
\usepackage{latexsym}

\usepackage[T1]{fontenc}

\usepackage[utf8]{inputenc}

\usepackage{microtype}

\usepackage{algorithm}
\usepackage[noend]{algpseudocode}
\usepackage{graphicx}
\usepackage{svg}
\usepackage{enumitem}
\usepackage{amsmath}
\usepackage{float}
\usepackage{caption}
\usepackage{subcaption}
\usepackage{soul}
\usepackage{multirow}
\usepackage{multicol}
\usepackage{numprint}
\usepackage{longtable}
\usepackage{booktabs, multirow}
\usepackage{changepage, threeparttable}
\usepackage{cuted}
\usepackage{makecell}
\usepackage{xcolor}
\usepackage{comment}
\usepackage{changepage}
\usepackage{amssymb}
\usepackage{pgfplots}
\pgfplotsset{width=0.45\textwidth,compat=1.9}

\graphicspath{ {./assets/} }

\title{UAlberta at SemEval-2023 Task 1: 
Context Augmentation and Translation \\
for Multilingual Visual Word Sense Disambiguation}

\author{Michael Ogezi, 
Bradley Hauer, 
Talgat Omarov, 
Ning Shi, 
Grzegorz Kondrak \\
  Alberta Machine Intelligence Institute \\
  Department of Computing Science \\ 
  University of Alberta, Edmonton, Canada \\
  \href{mailto:mikeogezi@ualberta.ca}{\color{black}\texttt{{\{mikeogezi,bmhauer,omarov,ning.shi,gkondrak\}@ualberta.ca}}} \\
}

\begin{document}

\maketitle

\begin{abstract}
We describe the systems of the University of Alberta team for the SemEval-2023 Visual Word Sense Disambiguation (V-WSD) Task. 
We present a novel algorithm that leverages glosses retrieved from BabelNet, in combination with text and image encoders. 
Furthermore, we compare language-specific encoders against the application of English encoders to translated texts.
As the contexts given in the task datasets are extremely short, we also experiment with augmenting these contexts with descriptions generated by a language model.
This yields substantial improvements in accuracy.
We describe and evaluate additional V-WSD methods which use image generation and text-conditioned image segmentation.
Overall, the results of our official submission rank us 18 out of 56 teams. 
Some of our unofficial results are even better than the official ones.
Our code is publicly available at {\url{https://github.com/UAlberta-NLP/v-wsd}}.
\end{abstract}

\section{Introduction} \label{intro}

This paper addresses our work on SemEval-2023 Task 1:  
Visual Word Sense Disambiguation\footnote{\url{https://raganato.github.io/vwsd/}} \cite{vwsd}.
The V-WSD task is closely related to 
WSD, and similarly involves understanding and classifying the meaning of a polysemous word in context. 
The distinction is in how classes are defined: 
In WSD, a system has access to a sense inventory that enumerates the possible senses of each word, 
and the task is to classify the focus word according to the sense that best corresponds to its intended meaning. 
In V-WSD, a system is given a set of candidate images,
and the task is to select the image which depicts the intended meaning of the focus word.

\begin{figure}[t]
  \centering
  {\includegraphics[width=0.1125\textwidth]{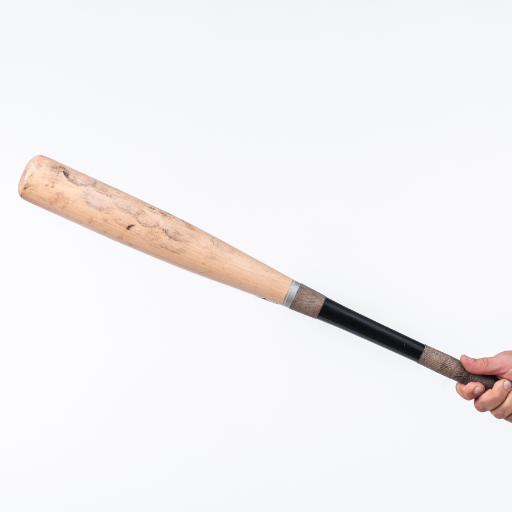}}
  {\includegraphics[width=0.1125\textwidth]{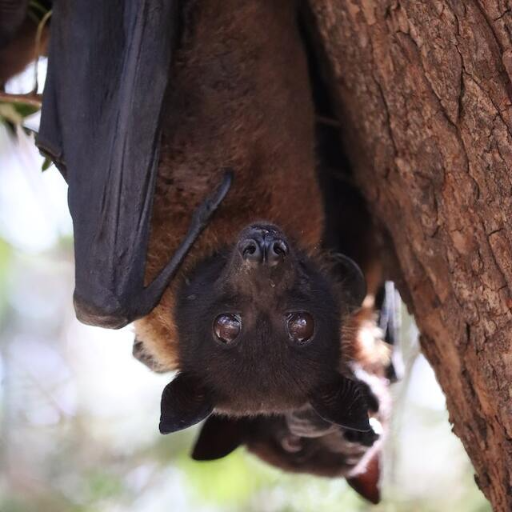}}
  {\includegraphics[width=0.1125\textwidth]{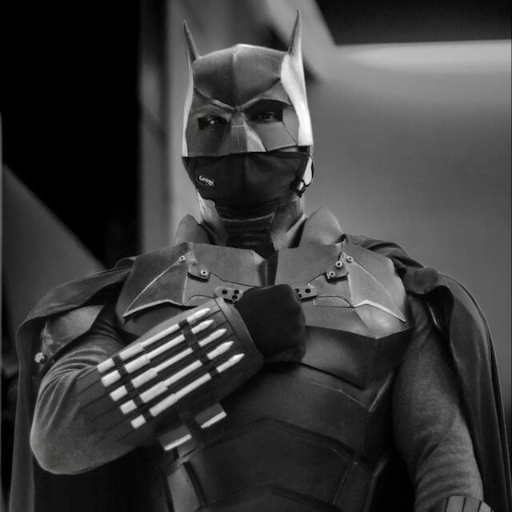}}
  {\includegraphics[width=0.1125\textwidth]{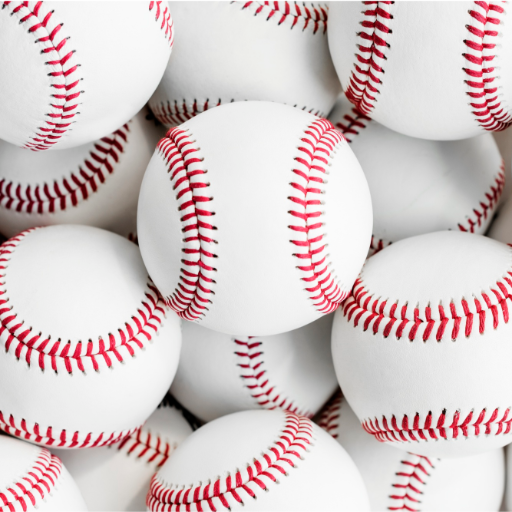}}
  
  \caption{
  The task is to select the image that best represents the meaning of the focus word (e.g., {\em bat}) in the context (e.g., {\rm ``baseball \underline{bat}}.'')}
  \label{figure:task_example}
\end{figure}

The multi-modal nature of V-WSD introduces challenges not encountered in WSD.
First, image processing is generally more computationally intensive than text processing.
Second, a V-WSD system must represent the meanings of both images and text, and must have mechanisms to compare these multi-modal semantic representations.
Last, since the candidate images in V-WSD are not restricted to a sense inventory, they may exhibit highly variable levels of sense granularity.

The V-WSD task is motivated by
cases where textual context alone 
is insufficient to disambiguate a word.
In such cases,
visual context may be available to facilitate disambiguation.
For example, the word \emph{play} is ambiguous in the context
``that was a good play,'' 
as it may refer to a theatrical performance
or an action in a sport.
However, an associated image of a stage or a sports field
would enable a V-WSD system to disambiguate \emph{play}.
%
%

We propose a novel V-WSD algorithm that ranks candidate images by embedding images and words-in-context in a shared semantic space, while also taking advantage of lexical knowledge bases commonly used in WSD. In particular, our method uses sense glosses of the focus word to create representations of the possible meanings that word may have. Our algorithm is flexible, 
It includes several optional modules, as well as hyper-parameters that facilitate customization, optimization, and detailed analysis. 

We test various configurations of our method and analyze their performance. 
Our three principal conclusions are as follows:
First and foremost, the augmention of the original textual context plays a crucial role in improving performance.
Second, there is a considerable gap between English and non-English performance,
indicating
that bias towards English models extends to the multi-modal setting.
Third, we observe a major distribution shift between the train and test sets, 
which is confirmed by our ablation study. 

\section{Related Work} \label{relwork}

Recent work on WSD can be divided into supervised and knowledge-based systems. 
Supervised WSD methods depend on large training corpora in which some or all of the content words have been tagged with their correct senses
\cite{blevins-zettlemoyer-2020-moving, consec}.
 %
Knowledge-based methods depend on other sources of linguistic knowledge
\cite{wang-wang-2020-synset}.
In general, knowledge-based methods are outperformed by contemporary supervised methods \cite{pasini2021}. Today, state-of-the-art WSD systems 
approach accuracy limits imposed by inter-annotator agreement \cite{maru2022}.


Research on the incorporation of visual information for WSD is relatively sparse. \citet{barnard-etal-2003-word} propose a statistical model that associates image regions and words to predict word senses. 
\citet{loeff-etal-2006-discriminating} apply spectral clustering to group similar images corresponding to the same senses. \citet{saenko-and-darrell-2008-unsupervised} employ an unsupervised approach to assign senses to images using surrounding texts 
and dictionary definitions, and then train a visual SVM classifier 
to disambiguate unseen images.
\citet{gella-etal-2019} introduce the task of visual verb sense disambiguation,
in which one image 
is selected based on a given context. \citet{vascon-etal-2021} propose a graph-based semi-supervised transductive
learning method for visual verb sense disambiguation.

Multi-modal foundation models \cite{foundation} such as CLIP \cite{clip} and ALIGN \cite{align} can represent both text and images in a shared embedding space.
Recent work\footnote{\url{https://github.com/moein-shariatnia/OpenAI-CLIP}} \cite{clip-italian, clip-fa} 
improves
the text encoder by using a pre-trained text-only encoder such as BERT \cite{bert}.

\section{Task \& Dataset} \label{task}

\label{par:definition}
\paragraph{Task Definition:} 
Given a focus word $w$ in a short context $c$,
and a set of candidate images $I$,
the task is to select the image $ i^* \in I$ 
which best represents the meaning of $w$ in $ c $. 
For example, given the context ``baseball \underline{bat}'' with {\em bat} as the focus word, 
a V-WSD system should choose 
the image that depicts the {\rm bat} used in baseball
(Figure~\ref{figure:task_example}). 

\paragraph{Dataset:}
The training data provided for this shared task consists of a silver dataset with 12,869 V-WSD instances.
Each sample is a 4-tuple $ \langle f, c, I, i^* \in I \rangle $ where $ |I| = 10 $.
The contexts are generally very short, often just a single word
in addition to the focus word.
We randomly select 10\% of the training data for use as a development set. 
The test dataset consists of 968 instances, of which 463 are English, 200 are Farsi, and 305 are Italian. 
We observe that many of the incorrect candidate images in the training data
have nothing to do with any sense of the focus word.
However, in the test data, we observe that this is less often the case, 
making the test set considerably more difficult.

\paragraph{Evaluation Metrics:} 
The primary metric is the hit rate, which is equivalent to top-1 accuracy,
or simply accuracy. 
This is the proportion of instances for which the system selects the correct image.
We also compute the mean reciprocal rank \cite{mrr} which represents how highly V-WSD systems rank the ground-truth image, on average.

\begin{table}[ht]
    \def\arraystretch{1.2}
    \centering
    \small
    \begin{tabular}{|c|c|l|}
    \hline
    {Language} & {\%} & Example \\
    \hline
    English & 82 & \underline{waxflower} wildflower \\
    Latin & 15 & \underline{shorea} genus \\
    German & 2 & \underline{truppenübungsplatz} workplace \\
    French & 1 & \underline{brumaire} month \\
    \hline
    \end{tabular}
    \caption{The language distribution of 100 instances 
    from the training set. The focus word is underlined. 
    }
    \label{table:language_distribution}
\end{table}

\paragraph{Language Distribution:} 
We observed some instances where the context contained non-English words.
To estimate the prevalence of this phenomenon, we randomly selected 100 instances from the training set and manually identified the language of each.
For example, the focus word \emph{shorea} in ``shorea genus'' comes from new Latin,
and refers to a genus of mainly rainforest trees. 
Table~\ref{table:language_distribution} shows the frequency of each language in our sample.

\section{Method} \label{method}

In this section, we describe the key components of 
our systems, including an algorithm that combines text and image similarity measures.

\subsection{Algorithm} \label{subs_alg}

We propose an algorithm to select a single image from a set of candidates that best matches the context. To reiterate the problem, we are given a context $c$ containing a focus word $w$ and a set $I$ of candidate images. 
We assume that we also have a non-empty set $G$ containing possible glosses of $w$; 
in practice, we obtain $G$ from BabelNet using the freely available API.\footnote{\url{https://babelnet.org/guide}}

Our algorithm makes calls to two similarity functions: 
The first is $sim^{L}$, 
a 
\emph{written language} 
similarity function, which takes as input two text strings and returns a value indicating the semantic similarity between them.
The second is $sim^{VL}$, 
a 
\emph{vision-to-written language} 
similarity function, which takes as input an image and a text string and returns a value indicating the similarity between 
what the image depicts
and 
what the text describes.

With these functions, for each candidate image $i \in I$, and for each gloss $g \in G$ of the focus word $w$, we compute the pairwise similarity between:
\begin{enumerate}
  \item The image and context: $s_{ic} = sim^{VL}(i, c)$
  \item The image and gloss: $s_{ig} = sim^{VL}(i, g)$
  \item The context and gloss: $s_{cg} = sim^{L}(c, g)$
\end{enumerate}
This allows us to
identify the pair of a candidate image $i^*$ and gloss $g^*$ that maximizes a weighted average of these three similarity scores.
%
Algorithm~\ref{algorithm} shows the pseudocode for this algorithm.
%

%

\begin{algorithm}[t]
    \normalsize
    \caption{Candidate Image Scoring}
    \begin{algorithmic}[1]
        \State $ c \gets $ the context of the focus word
        \State $ G  \gets $ list of glosses for the focus word 
        \State $ I \gets $ list of candidate images        
        \For{$ i $ in $ I $}
            \State $ s_{g} \gets 0 $ 
            \For{$ g $ in $ G $}
                \State $ s_{ig} \gets w_{ig} \cdot sim^{VL}({i, g}) $
                \State $ s_{cg} \gets w_{cg} \cdot sim^{L}({c, g}) $
                \State $ s_{g} \gets \max(s_{g}, s_{ig} + s_{cg}) $
            \EndFor
            \State $scores[i] \gets s_{g}
            + w_{ic} \cdot sim^{VL}({i, c}) $     
        \EndFor
        \State \Return $ scores $
    \end{algorithmic}
    \label{algorithm}
\end{algorithm}

    


\paragraph{Hyperparameters:} \label{paragraph:hyperparameters}
Our algorithm depends on three weight hyperparameters: $ w_{ic}$, $w_{ig}$, and $w_{cg}$. 
They represent the weights for image-context, image-gloss, and context-gloss similarity, respectively.
Table \ref{grid_search} shows the results of the hyperparameter binarized grid search performed on a 500-sample of the training set.
Based on our development experiment results, 
we decided to set all hyperparameter weights to $1$ for simplicity, 
except where otherwise noted. 
We discuss the hyperparameters further in Section \ref{hyperparameters}.

\begin{table}[bh]
\def\arraystretch{1.2}
\small
\centering
\begin{tabular}{|c|c|c|c|}
\hline
$w_{ic}$ & $w_{ig}$ & $w_{cg}$ & Accuracy (\%) \\
\hline
1 & 1 & 1 & 79.2 \\
1 & 1 & 0 & 79.2 \\
1 & 0 & 1 & 72.2 \\
1 & 0 & 0 & 72.2 \\
0 & 1 & 1 & 68.4 \\
0 & 1 & 0 & 68.6 \\
0 & 0 & 1 & 11.0 \\
\hline
\end{tabular}
\caption{Binarized grid search results for weight hyperparameters.}
\label{grid_search}
\end{table}


\paragraph{Context Augmentation:}
For each instance, we prompt InstructGPT \cite{gpt-3, instruct-gpt} to generate a definition for the context phrase. 
We use the following prompt template: 
{``For each line, define the phrase:''} 
followed by the contexts, one per line. 
For example,  
the context {``baseball bat''} 
is augmented to 
become {``baseball bat: a bat used to hit a baseball during the game of baseball.''}
The use of this additional context is described in Section \ref{sup}


\paragraph{Supplementary Training Data:}
We speculate that the size of the training dataset may be a limiting factor
in the accuracy of our method.
We, therefore, experiment with augmenting the training data
with additional data derived from BabelPic \citet{babelpic},
a multi-modal resource which maps a subset of BabelNet synsets to sets of one or more images.
For each pair of a synset and an image,
we enumerate a lemma from the base synset and a lemma from a related synset.
The two lemmas are concatenated, 
starting with the lemma from the related base synset,
to form a two-word context.
We then select nine other random images from BabelPic,
forming an instance comparable to those in the training set:
a two-word context with a single focus word,
with ten images, one depicting the correct sense of the focus.
We create 54,968 instances this way and experiment with adding this dataset
to the training data at training time.

\paragraph{Glosses:}
For each instance, we enumerate the BabelNet \cite{babelnet} glosses corresponding to each sense of the focus word. If there are multiple glosses for a single sense, we pick the first and add it to the set $G$. This prevents senses from being over-represented 
due to the number of glosses in BabelNet.

\section{Systems}

In this section, we describe our systems for the V-WSD task, 
Our official system submissions are based on our primary systems: 
\textsc{Tr} and \textsc{LangSpec}.
We also describe two alternative systems, which do not use Algorithm~\ref{algorithm}. 
Both perform worse than the primary systems, 
but their results are nevertheless valuable for the purpose of analysis.
We also present a supplementary method, 
which can be optionally used in combination with our other systems.  
Non-English instances are translated using DeepL\footnote{\url{https://www.deepl.com/translator}} for Italian and ChatGPT\footnote{\url{https://openai.com/blog/chatgpt/}} for Farsi. 
%

\subsection{Primary Systems}

\paragraph{\textsc{Tr}: Image Scoring with Translations}
If the input instance is not English, we translate it into English.
Then we apply Algorithm~\ref{algorithm}.
We compute $ sim^{VL} $ using embeddings from CLIP \cite{clip}, an English-only model which encodes text and images in a shared embedding space. We compute $ sim^{L} $ using BERT \cite{bert} as an English-only text encoder. We set the weight parameters: $w_{ic}$, $w_{ig}$, and $w_{cg}$ to 1 in this specific case.

\paragraph{\textsc{LangSpec}: Image Scoring with Language-Specific Models}
This system is similar to \textsc{Tr}, except that non-English instances are not translated into English. 
This is our only system which directly operates in other languages. 
Given a non-English instance, we replace CLIP and BERT with language-specific models to compute $sim^{VL}$ and $sim^{L}$. For English instances, this method is the same as \textsc{Tr}. For Italian, we use CLIP-Italian \cite{clip-italian} to compute $sim^{VL}$ and Italian BERT\footnote{\url{https://huggingface.co/dbmdz/bert-base-italian-xxl-uncased}} to compute $sim^{L}$. For Farsi, we use CLIPfa \cite{clip-fa} to compute $ sim^{VL} $ and ParsBERT \cite{pars-bert} to compute $ sim^{L} $. 

\subsection{Alternative Systems}

\paragraph{\textsc{Gen}: Generative Image Model}
This method takes a different approach compared to \textsc{Tr} and \textsc{LangSpec}; it does not use Algorithm~\ref{algorithm}.
Instead, we provide the context (translated into English, if needed, as outlined above) as input to Stable Diffusion \cite{stable}, a generative model which takes a text prompt as input and produces candidate images to depict what the text describes. For each context, we generate $15$ images using $20$ diffusion steps each. We set the guidance scale hyperparameter to $7.5$.
For each candidate image, we compute its cosine similarity with each generated image based on embeddings produced by CLIP. The candidate with the highest similarity to the generated images is chosen as the output.

\paragraph{\textsc{Seg}: Text-Conditioned Image Segmentation}
As with \textsc{Gen}, this method does not use Algorithm~\ref{algorithm}. Instead, we use a zero-shot image segmentation system \cite{clipseg} to segment images based on the provided context. This system produces a \emph{mean mask value}, which we use as a measure of similarity between the context and the segmented image; we return the image with the highest mean mask value, given the context.

\subsection{Supplementary Method}
\label{sup}
\paragraph{\textsc{Def}: Generating Additional Context} \label{paragraph:x_def}
\textsc{Tr}, \textsc{Gen}, and \textsc{Seg} make use of the input context, translated to English as needed. However, the contexts provided in the official dataset for this task are extremely short. With \textsc{Def}, we generate additional context by using the original context to prompt InstructGPT for a more extensive description,
as described in Section \ref{subs_alg}. 
We then concatenate the generated text to the context and pass this augmented context to \textsc{Tr}, \textsc{Gen}, or \textsc{Seg}. 
We refer to the methods using this supplementary method as
\textsc{Tr+Def}, \textsc{Gen+Def}, and \textsc{Seg+Def}, respectively. 
We do not combine \textsc{Def} with \textsc{LangSpec}, as we observe that InstructGPT is less robust to short non-English contexts. 

For \textsc{Tr+Def}, we set $w_{ig}$ and $w_{cg}$ to 0, as the improved context obviates the need for their corresponding terms
in Algorithm~\ref{algorithm}. 
\textsc{Gen+Def} and \textsc{Seg+Def}, 
being based on \textsc{Gen} and \textsc{Seg}, 
do not depend on Algorithm~\ref{algorithm}.

\section{Experiments}

In this section, we present, discuss, and analyze our results.

\newcommand{\bb}[1]{\textbf{#1}}

\begin{table}[t]
    \centering
    \begin{tabular}{lcccc}
    \hline
    & EN            & IT        & FA        & Avg \\ 
    \hline
    Baseline                         & 60.5          & 22.6      & 28.5      & 37.2      \\
    \textsc{Tr*}                      & 61.1          & 59.3      & \bb{43.0}      & 54.5      \\
    \textsc{Tr+Def}                  & \bb{69.1}     & \bb{63.3} & 40.0 & \bb{57.5} \\
    \textsc{LangSpec*}                & 56.8          & 37.7      & 14.5      & 36.3      \\
    \textsc{Gen}                     & 51.6          & 45.9      & 39.0         & 45.5         \\
    \textsc{Gen+Def}                 & 58.1          & 48.5      & 34.5         & 47.0         \\
    \textsc{Seg}                  & 31.5          & 29.8      & 20.5      & 27.3         \\
    \textsc{Seg+Def}                 & 34.1          & 36.7      & 20.0      & 30.3         \\
    \hline
    \end{tabular}    
    \caption{
    Accuracy for English, Italian, and Farsi, along with the macro average for all languages. 
    We indicate our official system submissions with *.
    }
    \label{table:main_results}
\end{table}

\subsection{Results} 
\label{results}

Table \ref{table:main_results} shows our performance on the test set.
We find that accuracy has a 99.46\% Pearson correlation with mean reciprocal rank, and so for conciseness, we report accuracy alone.
The translation-based systems, \textsc{Tr} and \textsc{Tr+Def},
yield the best results. 
One explanation for this outcome is
the disproportionate amount of English training data 
available to the models we build upon: CLIP and BERT.
The higher performance of these models on English
appears to compensate for the noise introduced by the translation process.
We discuss this further in Section \ref{subsection:english_lang_hegemony}.

An interesting trend 
is the benefit of context augmentation,
(Section \ref{paragraph:x_def}). 
Between \textsc{Tr} and \textsc{Tr+Def}, 
we observe a 3\% average improvement in accuracy. 
We observe a similar trend in \textsc{Gen} versus \textsc{Gen+Def} 
and \textsc{Seg} versus \textsc{Seg+Def}.

We further observe 
that accuracy on English instances is highest, 
accuracy on Farsi instances is lowest, 
while accuracy on Italian instances is in between both. 
This corresponds to the quality and quantity of resources available 
for each language. 
We undertake more thorough analyses in the next section.


\subsection{Distribution Shift}

As shown in Table \ref{table:distribution_shift}, we observe a clear disparity in polysemy, 
and the proportion of focus words which are nouns, 
between the training and test sets. 
This difference is especially notable when considering the performance gap between the sets.

\begin{table}[ht]
\centering
\begin{tabular}{l|c|ccc}
\hline
 & {Train} & \multicolumn{3}{c}{Test} \\
\cline{2-5}
 & EN & EN & IT & FA \\
\hline
{Polysemy} & 6.8 & 23.1 & 13.6 & 10.7 \\
Nouns (\%) & 74.7 & 88.1 & 91.5 & 92.5 \\
\hline
\end{tabular}
\caption{Distribution shifts between the training and test sets. {Polysemy} indicates the average number of senses each focus word in the set has.}
\label{table:distribution_shift}
\end{table}

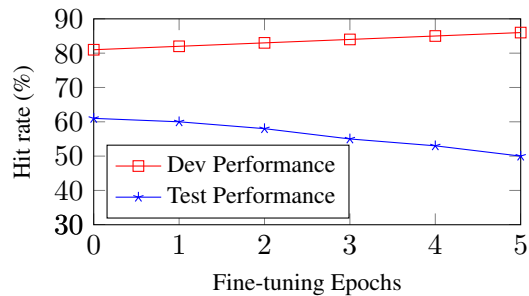
\begin{figure}[ht]
\begin{adjustwidth}{}{-8em}
\begin{tikzpicture}
\begin{axis}[
    height=0.4\linewidth,
    xlabel={\small Fine-tuning Epochs},
    ylabel={\small Hit rate (\%)},
    xmin=0, 
    xmax=5,
    ymin=30, 
    ymax=90,
    xtick={0,1,2,3,4,5},
    ytick={30,30,40,50,60,70,80,90},
    legend pos=south west,
    grid style=dashed,
]

\addplot[color=red, mark=square]
    coordinates {(0,81)(1,82)(2,83)(3,84)(4,85)(5,86)};
    \addlegendentry{\small Dev Performance}

\addplot[color=blue, mark=star]
    coordinates {(0,61)(1,60)(2,58)(3,55)(4,53)(5,50)};
    \addlegendentry{\small Test Performance}
    
\end{axis}
\end{tikzpicture}
\end{adjustwidth}
\caption{
As we fine-tune on the training set for more epochs, we see an increase in dev set performance, but a drop in test set performance.
Epoch 0 refers to using the model zero-shot, with no fine-tuning.
}
\label{figure:fine-tuning_penalty}
\end{figure}

\paragraph{Zero-shot vs. Fine-tuning:}
We observe that fine-tuning on the training set 
leads to a drop in performance on the test set
(Figure \ref{figure:fine-tuning_penalty}).
This may be due to the divergence between the training and test datasets
outlined above.

\subsection{Traditional WSD}

Although both the V-WSD and WSD tasks have some similarities, 
we found that some ideas drawn from WSD prove ineffective for V-WSD.

\paragraph{Using Glosses:} 
We observe empirically worse performance when using glosses in our algorithm. Specifically, with \textsc{Tr} on the English test set, we obtain a hit rate of 61.1\% when we do not use glosses and 56.8\% when we do. Such a steep drop (4.3\%) is surprising,
especially since most state-of-the-art WSD systems explicitly use glosses in their methods.

We posit that sense disambiguation in V-WSD is more focused on homonymy than polysemy and, as a result, can be less nuanced than in WSD. For example, \textit{apple} could refer to a fruit or a tree. In an image depicting both, the focus may be unclear. In WSD, this distinction is critical since \textit{tree} and \textit{fruit} are distinct senses. In V-WSD, however, we can make a correct prediction without deciding between both senses. As a result of this lower granularity, glosses become less important.

\paragraph{Performance of WSD Systems on Context:}
We manually disambiguated the sense of the focus word 
in a randomly-selected set of 16 instances from the training set.
We then applied a state-of-the-art WSD system, ConSec \cite{consec}, to these instances. 
We observe that ConSec sense predictions were accurate 50\% of the time,
falling considerably below its reported accuracy of 82\%.



\subsection{English 
Hegemony} 
\label{subsection:english_lang_hegemony}

Natural language processing research often focuses on the English language,
at the expense of other languages \cite{low-resource}.
The relative performance of \textsc{Tr} and \textsc{LangSpec} reflects this phenomenon:
Translating non-English text to English,
in order to apply an English encoder,
can be expected to introduce some noise due to translation errors and information loss.
However, we observe that this pipeline approach produces better results
than using an Italian or Farsi encoder directly.
This suggests that the field's focus on English has yielded English encoders
which are much better than those available for Italian or Farsi.
We speculate that advancing the state-of-the-art for non-English encoders
may yield even better performance, by avoiding the need to translate to English.
%


\subsection{Image Generation}

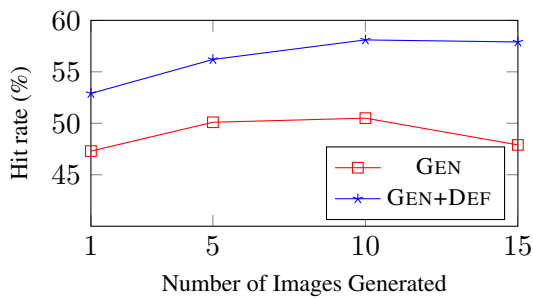
\begin{figure}[ht]
\begin{adjustwidth}{}{-8em}
\begin{tikzpicture}
\begin{axis}[
    height=0.4\linewidth,
    xlabel={\small Number of Images Generated},
    ylabel={\small Hit rate (\%)},
    xmin=1, 
    xmax=15,
    ymin=40, 
    ymax=60,
    xtick={1,5,10,15},
    ytick={45,50,55,60},
    legend pos=south east,
    grid style=dashed,
]
\addplot[color=red, mark=square]
    coordinates {(1,47.3)(5,50.1)(10,50.5)(15,47.9)};
    \addlegendentry{\textsc{\small Gen}}
\addplot[color=blue, mark=star]
    coordinates {(1,52.9)(5,56.2)(10,58.1)(15,57.9)};
    \addlegendentry{\textsc{\small Gen+Def}}
\end{axis}
\end{tikzpicture}
\end{adjustwidth}
\caption{Hit rate (\%) vs. number of images generated for \textsc{Gen} and \textsc{Gen+Def}.}
\label{figure:number_of_images_generated}
\end{figure}

As shown in Figure \ref{figure:number_of_images_generated}, when applying our image generation system (\textsc{Gen}), we observe an increase in performance as we generate more images. Although the performance jump when transitioning from 1 to 5 images is most pronounced, we see benefits from scaling until a certain point, 10 images, where the trend becomes unreliable.

\subsection{Text-Conditioned Image Segmentation}

With \textsc{Seg}, we can sometimes robustly segment images and predict masks indicating the correct image, conditioning on the full context. However, this method sometimes forms incorrect semantic representations. Appendix \ref{appendix:full_segmentation} details more examples of \textsc{Seg}'s usage.
In addition to Figure \ref{figure:segmentation}, we present more extensive examples in Appendix \ref{appendix:full_segmentation}.

\begin{figure}[t]
  \centering
  {\includegraphics[width=0.1125\textwidth]{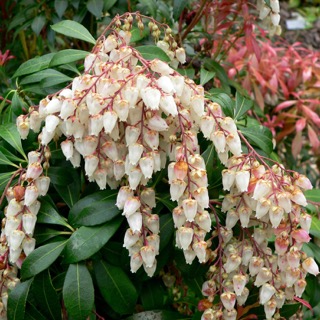}}
  {\includegraphics[width=0.1125\textwidth]{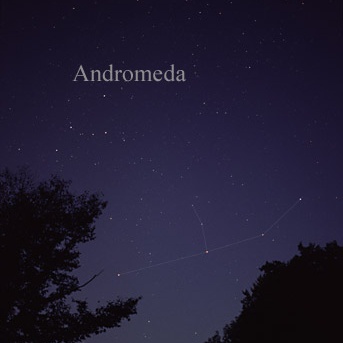}}
  {\includegraphics[width=0.1125\textwidth]{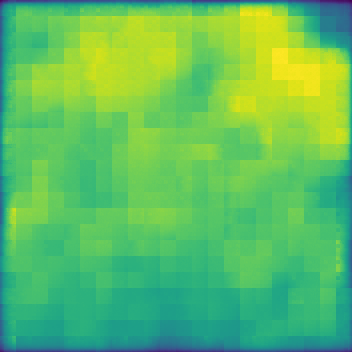}}
  {\includegraphics[width=0.1125\textwidth]{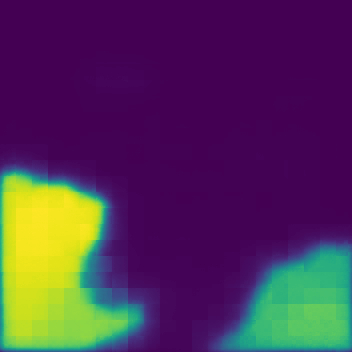}}
  \caption{
  Original images from the dataset depicting \textsc{\color{blue}{andromeda}} (Japanese plant), \textsc{\color{red}{andromeda}} (galaxy) and their two masks conditioned on {{``{andromeda} tree.''}}
  }
  \label{figure:segmentation}
\end{figure}

\label{hyperparameters}
\subsection{Algorithm Hyperparameters}
Our algorithm uses three weights hyperparameters to balance pairwise similarities. We set all weights to 1 based on Table \ref{grid_search}.
Comparison of results with $w_{cg}$ set to 0 or 1 suggests that $sim^{L}(c, g)$ does not improve performance. Two reasons support this finding.
Firstly, images encode richer representations, producing more precise $sim^{VL}(i, g)$ and $sim^{VL}(i, c)$, while both context and glosses are discrete textual features, introducing uncertainty to $sim^{L}(c, g)$.
Secondly, we use CLIP and BERT to calculate $sim^{VL}$ and $sim^{L}$, respectively. CLIP's multi-modal pre-training may offer better similarity scores, fitting this task better. Understanding these findings more deeply is an interesting avenue for future research.

\section{Conclusion} 
\label{conclusion}

In this paper, we outlined our work on the recently proposed task of V-WSD.
We found that many ideas from traditional WSD 
are difficult to adapt to V-WSD, and, moreover, WSD systems are generally not useful for V-WSD.
%
We were particularly surprised to find that, unlike in WSD, glosses appear to be unhelpful for V-WSD. 
Contrariwise, our innovation of 
augmenting the context did yield substantial gains in accuracy.

Further research will be needed to establish the connection between V-WSD and the broader field of lexical semantics.
We speculate that developing systems for joint WSD and V-WSD may yield improvements in one or both tasks.
%
Our work here serves as a proof-of-concept 
establishing the utility of language models and lexico-semantic resources 
in the developing task of V-WSD.

\section*{Acknowledgements}


This research was supported by the Natural Sciences and Engineering Research Council of Canada (NSERC), and the Alberta Machine Intelligence Institute (Amii).
We also thank Behzad Shayegh for providing advice on the Farsi test set. 

\bibliography{anthology,custom}
\bibliographystyle{acl_natbib}


\appendix
\clearpage

\section{Text-Conditioned Image Segmentation}
\label{appendix:full_segmentation}

\subsection{Success Mode}
In the successful case of this system, we see that we are able to segment the object based on the text provided properly.
See the figures below for details.

\begin{figure}[ht!]
  \centering
  {\includegraphics[width=0.225\textwidth]{assets/172.jpeg}}
  \hfill
  {\includegraphics[width=0.225\textwidth]{assets/173.jpg}}
  
  \caption{Original images from the dataset depicting on the left: \textsc{\color{blue}{andromeda}} and on the right: \textsc{\color{red}{andromeda}}.}
  \vspace{1em}
  
  {\includegraphics[width=0.225\textwidth]{assets/andromeda_tree_172.png}}
  \hfill
  {\includegraphics[width=0.225\textwidth]{assets/andromeda_tree_173.png}}

  \caption{Conditioned on the full {{``{andromeda} tree''}}}
  \vspace{1em}
  
  {\includegraphics[width=0.225\textwidth]{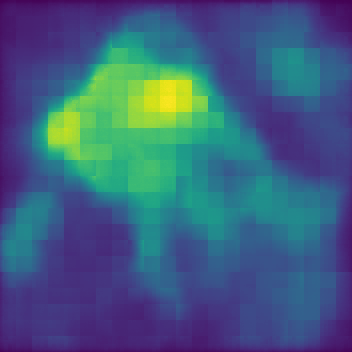}}
  \hfill
  {\includegraphics[width=0.225\textwidth]{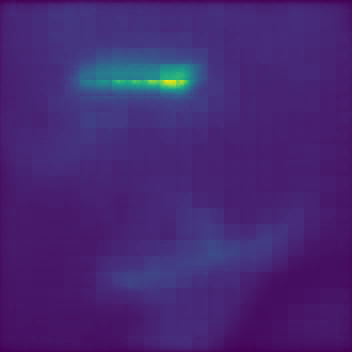}}

  \caption{Conditioned on {``andromeda''}}
  \vspace{1em}
  
  {\includegraphics[width=0.225\textwidth]{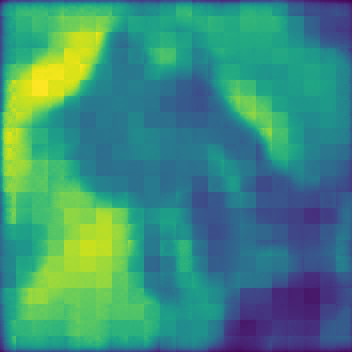}}
  \hfill
  {\includegraphics[width=0.225\textwidth]{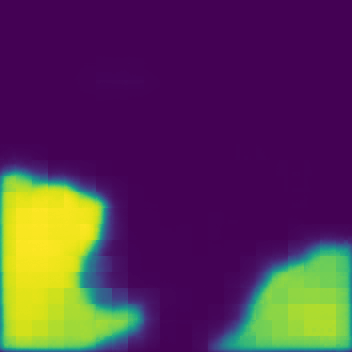}}
  
  \caption{Conditioned on {``tree''}}
  \label{figure:full_segmentation}
\end{figure}

\newpage
\subsection{Failure Mode}
In the failure case of this system, we see that we are unable to confidently segment the object based on the text provided. 
See the figures below for details.

\begin{figure}[ht!]
  \centering
  {\includegraphics[width=0.225\textwidth]{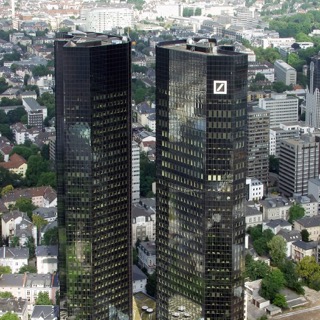}}
  \hfill
  {\includegraphics[width=0.225\textwidth]{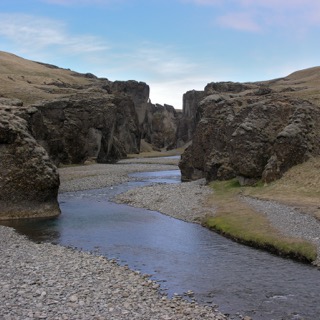}}
  
  \caption{Original images from the dataset depicting on the left: \textsc{\color{blue}{bank}} (finance) and on the right: \textsc{\color{red}{bank}} (river).}
  \vspace{1em}
  
  {\includegraphics[width=0.225\textwidth]{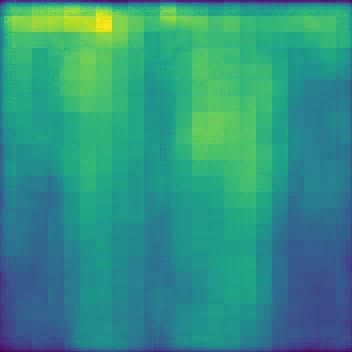}}
  \hfill
  {\includegraphics[width=0.225\textwidth]{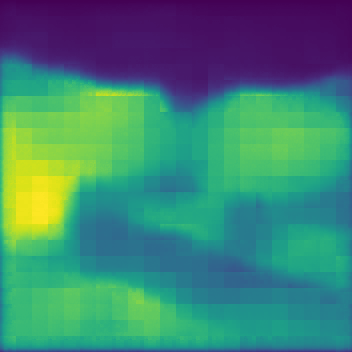}}

  \caption{Conditioned on the full {``{bank} erosion''}}
  \vspace{1em}
  
  {\includegraphics[width=0.225\textwidth]{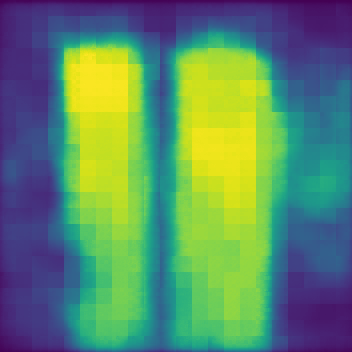}}
  \hfill
  {\includegraphics[width=0.225\textwidth]{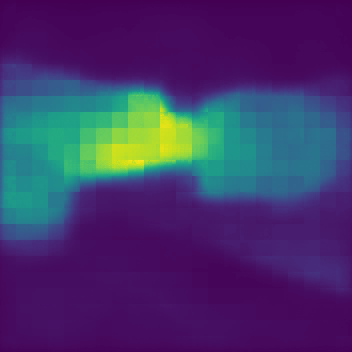}}

  \caption{Conditioned on {``bank''}}
  \vspace{1em}
  
  {\includegraphics[width=0.225\textwidth]{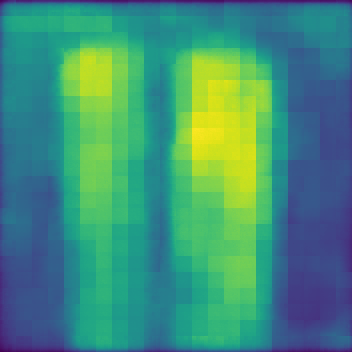}}
  \hfill
  {\includegraphics[width=0.225\textwidth]{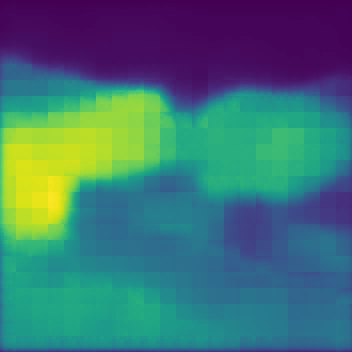}}
  
  \caption{Conditioned on {``erosion''}}
  \label{figure:full_segmentation_failure}
\end{figure}


\end{document}